\documentclass[10pt,twocolumn,letterpaper]{article}
\pdfoutput=1
\usepackage[final]{cvpr}
\usepackage{times}
\usepackage{graphicx}
\usepackage{amsmath}
\usepackage{amssymb}
\usepackage{booktabs}
\usepackage{multirow}
\usepackage{xcolor}
\usepackage[hidelinks]{hyperref}
\hypersetup{
  pdftitle={Mind the Gap: Standard 3DGS Evaluation Primarily Measures Near-Trajectory Interpolation},
  pdfauthor={Gaoxiang Jia, Vikram Appia}
}
\usepackage{subcaption}
\usepackage{enumitem}

\title{Mind the Gap: Standard 3DGS Evaluation Primarily Measures\\Near-Trajectory Interpolation}

\author{
Gaoxiang Jia$^{*}$\quad Vikram Appia\\[3pt]
Advanced Micro Devices, Inc.\\[3pt]
{\small $^{*}$Corresponding author: \texttt{jiagaoxiang@gmail.com}}
}

\begin{document}
\maketitle

\begin{abstract}
Standard MipNeRF360-style 3D Gaussian Splatting (3DGS) evaluation holds
out every $N$-th frame---but these frames have trained neighbors on
both sides, so the metric measures \emph{near-trajectory interpolation}
rather than spatial generalization. We introduce a \emph{fair matched-count}
protocol that isolates this effect: both arms train on the same number
of images and differ only in whether the holdout is spread evenly
(interpolation) or forms a contiguous spatial sector (extrapolation).
Our primary finding is a large, consistent interpolation--extrapolation
gap of \textbf{3--12~dB}---several times the differences typically
reported between competing methods. The gap is robust to training
noise, is in two cases large enough to flip a method ranking under
multi-seed confirmation, and---crucially---persists across three
representation families, including a non-Gaussian volumetric neural
radiance field (NeRF), so it reflects spatial coverage rather than any
one representation.
Diagnostically, it is dominated by a diffuse/geometry-proxy component
and tracks each view's angular distance to its nearest training view, a
zero-cost signal that also guides capture planning; loss-side
regularization yields only marginal gains.
Standard holdouts remain useful for near-trajectory rendering but
should not, alone, be read as evidence of spatial generalization.
Prior work notes protocol sensitivity; ours is, to our knowledge, the
first to combine matched-count paired holdout, cross-representation
quantification, and a diagnostic analysis (Table~\ref{tab:prior}). We
describe a \emph{spatial-holdout benchmark toolkit} with standardized
splits and baselines for 16 scenes, which we are preparing for public
release.
\end{abstract}

\section{Introduction}
\label{sec:intro}

3D Gaussian Splatting (3DGS)~\cite{kerbl3Dgaussians} has rapidly
become the preferred representation for real-time novel-view
synthesis~\cite{nerf}, owing to its explicit scene representation and
efficient rasterization.
Building on this foundation, recent \emph{world-model} pipelines
generate panoramic environments and then reconstruct them as 3DGS
scenes using video-diffusion models to produce multi-view training
frames~\cite{hyworld,wonderworld,cat3d}.

A critical question underlies all such systems: \emph{how do we know
the 3D reconstruction is any good?} The standard answer is peak
signal-to-noise ratio (PSNR), structural similarity (SSIM), and
learned perceptual image patch similarity (LPIPS) computed on held-out
views selected by subsampling every
$N$-th frame from an ordered camera sequence. This
\texttt{test\_every} convention, used since the original
3DGS~\cite{kerbl3Dgaussians}, has a structural limitation: each held-out
frame has trained neighbors on both sides, so the metric primarily
measures how well the model \emph{interpolates} between nearby trained
views rather than how well it generalizes to unseen spatial regions.

\paragraph{Why extrapolation matters.}
Many applications require \emph{spatial generalization}, not just
near-trajectory rendering: driving simulators render adjacent-lane
views~\cite{euvs,vegs}, virtual and augmented reality (VR/AR) lets
users move beyond the captured path,
sparse-capture (real estate, e-commerce) synthesizes from few photos,
and world-model pipelines generate training data from novel
viewpoints~\cite{hyworld,wonderworld}. Yet even for near-capture
rendering the standard metric gives \emph{false confidence}: a model
scoring 30+~dB on interpolation can drop to 18--25~dB on spatial
holdout (Table~\ref{tab:main}), with no warning. A benchmark that
claims ``novel-view synthesis'' but tests near-neighbor interpolation
can inadvertently steer method selection toward reproducing training
views rather than generalizing.

Prior work has observed aspects of this
problem~\cite{nerfbusters,euvs,vegs,nerfbaselines,freevs,nerfdirector}.
However, no existing study combines these axes (Table~\ref{tab:prior}):
(1)~matched training-set size, (2)~a dB gap across multiple methods and
scenes, (3)~multi-seed-confirmed ranking changes under that protocol,
(4)~a spherical harmonic (SH) degree analysis, and (5)~generality
across representation families---feed-forward Gaussians \emph{and} a
non-Gaussian volumetric NeRF.
In short: prior work observes protocol sensitivity; we provide the
first matched-count, multi-representation benchmark with a diagnostic
analysis, showing the effect is \emph{larger than method deltas}.

We propose a \textbf{fair matched-count comparison}: hold out the same
number of images either spread evenly (interpolation, the standard
protocol) or as a contiguous angular sector (extrapolation). Both arms
train on identical data counts and evaluate on genuinely unseen
images---the designed contrast is the spatial distribution of the
holdout. Across 16 scenes (10 real-capture, 6 generated-view) and three
published 3DGS implementations, we consistently measure gaps of
\textbf{3--12~dB}---larger than the performance differences typically
reported between published methods (Table~\ref{tab:main}: inter-method
range $<$2.5~dB on interpolation). A method comparison
claiming ``0.5~dB better'' is making a claim smaller than the
gap between evaluation modes.

Our contributions:
\begin{enumerate}[leftmargin=*,topsep=2pt,itemsep=1pt]
\item A \textbf{fair matched-count evaluation protocol} that measures
  interpolation and extrapolation on genuinely unseen images with
  identical training data counts (\S\ref{sec:method}).
\item A \textbf{systematic quantification} of the gap across 16
  scenes, 3 methods, and 502 training runs with targeted multi-seed
  validation---showing the gap is 39$\times$ larger than training
  noise and is, in two multi-seed-confirmed cases, consequential for
  method ranking (\S\ref{sec:gap}).
\item \textbf{Mechanistic decomposition}: SH-degree analysis
  attributes 62\% of the gap to a diffuse/geometry-proxy component and
  38\% to view-dependent color, with
  higher SH degrees \emph{hurting} extrapolation on 3/9 scenes
  (\S\ref{sec:mechanism}). Per-image quality correlates with
  nearest-training-view distance ($r\!=\!-0.58$, $n\!=\!492$),
  providing a zero-cost diagnostic.
\item \textbf{Coverage-guided view selection}: a zero-cost heuristic
  that outperforms random placement on 31/40 scene-budget combinations
  (10 scenes, 240 runs), strongest at $M\!=\!1$ (9/10 scenes).
\item \textbf{Cross-representation generality}: under our matched-count
  protocol the gap persists in a non-Gaussian NeRF (Instant-NGP;
  +3.7~dB, $r\!=\!0.90$ with the 3DGS gap), and an analogous gap appears
  in two feed-forward methods under a distance-based target protocol
  (MVSplat, DepthSplat; +7--9~dB in-domain on RealEstate10K)---the
  effect tracks spatial coverage, not Gaussian primitives
  (\S\ref{sec:mechanism},~\S\ref{sec:discussion}).
\item A \textbf{spatial-holdout benchmark toolkit}: split generation,
  a one-command evaluator, and a pose-only coverage diagnostic, with
  standardized splits and baseline metrics for 16 scenes
  (Appendix~\ref{sec:supp_toolkit}).\footnote{The benchmark toolkit and accompanying code
  are still in preparation for release; we plan to make them publicly
  available in a future revision of this paper. In the meantime, the
  protocol, split-construction rule, and per-scene configurations are
  documented in full in the appendix so the results can be
  reproduced independently.}
\end{enumerate}

\section{Related Work}
\label{sec:related}

\paragraph{3D Gaussian Splatting.}
3DGS~\cite{kerbl3Dgaussians} represents scenes as collections of
anisotropic 3D Gaussians with learned positions, covariances,
opacities, and spherical harmonic coefficients. Subsequent work has
improved densification~\cite{absgs,3dgsmcmc}, introduced
structured anchors~\cite{scaffoldgs}, pursued
compression~\cite{compact3dgs,minisplatting}, anti-aliasing~\cite{mipgs},
2D formulations~\cite{2dgs}, progressive propagation~\cite{gaussianpro},
principled pruning~\cite{popgs}, efficient densification~\cite{edgs},
and appearance embedding~\cite{nerfw,hanerf}. Feed-forward methods predict
Gaussians directly from sparse images~\cite{pixelsplat,mvsplat,freesplat,instantsplat},
bypassing per-scene optimization but inheriting training-distribution
coverage assumptions. Nearly all of these report results under the standard
Mip-NeRF360 protocol, commonly using every 8th ordered image as a
test view; our experiments show that this sequential-holdout setting
can be systematically biased when the ordered images lie on smooth
camera trajectories.

\paragraph{Generated-View 3DGS.}
World-model pipelines generate multi-view training data from
panoramic images using video-diffusion
models~\cite{hyworld,wonderworld,cat3d,viewcrafter}. Related systems
include LucidDreamer~\cite{luciddreamer},
ZeroNVS~\cite{zeronvs}, and WildGaussians~\cite{3dgswild}
for in-the-wild captures. These systems use a mix of evaluation
protocols, including held-out target views, fixed render paths, and
within-trajectory splits. Our critique applies specifically when
evaluation holds out frames along the same smooth generated trajectory:
the held-out frames are then interpolation on the generated path rather
than spatial extrapolation.

\paragraph{Evaluation of Novel-View Synthesis.}
A growing body of work identifies limitations in standard novel-view
synthesis (NVS) evaluation. Several propose separate train/test
trajectories or
extrapolation benchmarks~\cite{nerfbusters,euvs,vegs,viewextrapolator,extranerf,slamrender,geoevs,extrags},
and others show evaluation protocol can change or even invert reported
rankings~\cite{nerfbaselines,nerfdirector,3dgsIeval}. Closest to us,
FreeVS~\cite{freevs} independently observes that periodic frame holdout
is ``novel frame synthesis'' rather than novel-view synthesis and
proposes dropping camera viewpoints---a convergent diagnosis that
underscores the problem's importance.
Our work differs in three ways that, to our knowledge, no prior study
combines (Table~\ref{tab:prior}): a \emph{matched-count} protocol
(both arms train on $N\!-\!K$ images, removing the data-count confound
that makes prior comparisons ambiguous); a \emph{quantified} dB gap
across methods, scenes, and representation families; and a
\emph{diagnostic} analysis (SH decomposition, angular-distance
correlation) rather than an observation alone.

\begin{table}[t]
\centering
\setlength{\tabcolsep}{2pt}
\footnotesize
\caption{\textbf{Comparison with prior evaluation studies}
along axes we combine; entries reflect each work's \emph{primary
reported protocol}, not a judgment of quality.
\emph{Matched count}: interp and extrap arms train on the same number
of images. \emph{Paired interp/ext}: both modes measured on the same
scenes under one protocol. \emph{Multi-method}: $\geq$2 reconstruction
methods compared. To our knowledge ours is the first to combine all
listed axes.}
\label{tab:prior}
\begin{tabular}{lccccccc}
\toprule
& \rotatebox{70}{Matched count}
& \rotatebox{70}{Paired interp/ext}
& \rotatebox{70}{Multi-method}
& \rotatebox{70}{Real+generated}
& \rotatebox{70}{Multi-seed}
& \rotatebox{70}{SH analysis}
& \rotatebox{70}{Toolkit} \\
\midrule
Nerfbusters~\cite{nerfbusters} & & & & & & & \\
NeRF Dir.~\cite{nerfdirector} & & \checkmark & & & & & \\
EUVS~\cite{euvs} & & & \checkmark & & & & \\
VEGS~\cite{vegs} & & & & & & & \\
FreeVS~\cite{freevs} & & & \checkmark & & & & \\
NerfBase.~\cite{nerfbaselines} & & & \checkmark & & & & \checkmark \\
\textbf{Ours} & \checkmark & \checkmark & \checkmark & \checkmark & \checkmark & \checkmark & \checkmark \\
\bottomrule
\end{tabular}
\end{table}

\paragraph{Regularization for 3DGS Generalization.}
RegNeRF~\cite{regnerf} pioneered regularization from unseen viewpoints
for few-shot NeRF; several 3DGS works build on this principle.
ICO-GS~\cite{icogs} identifies ``appearance compensation''---the
optimizer adjusting color/opacity to mask geometric errors---as a key
failure mode under sparse views.
DropoutGS~\cite{dropoutgs} and the concurrent
DropGaussian~\cite{dropgaussian} independently propose randomly
dropping Gaussians during training for sparse-view robustness.
CoherentGS~\cite{flatminima} addresses sparse-view overfitting by
seeking flatter loss landscapes, while FASR~\cite{fasr} extends
this with frequency-adaptive sharpness-aware minimization.
Co-Adaptation~\cite{coadaptation} formalizes why Gaussians become
entangled to fit training views and provides theoretical justification
for dropout-based regularization.
DN-Splatter~\cite{dngsplat} incorporates depth and normal priors
for improved geometry.
Our work extends these insights to the evaluation-gap setting:
we test Gaussian dropout as a view-specialization regularizer and
find its effect is scene-dependent---it does not universally reduce
the gap, suggesting that the gap is largely a data-coverage problem.

\begin{figure}[t]
\centering
\includegraphics[width=\columnwidth]{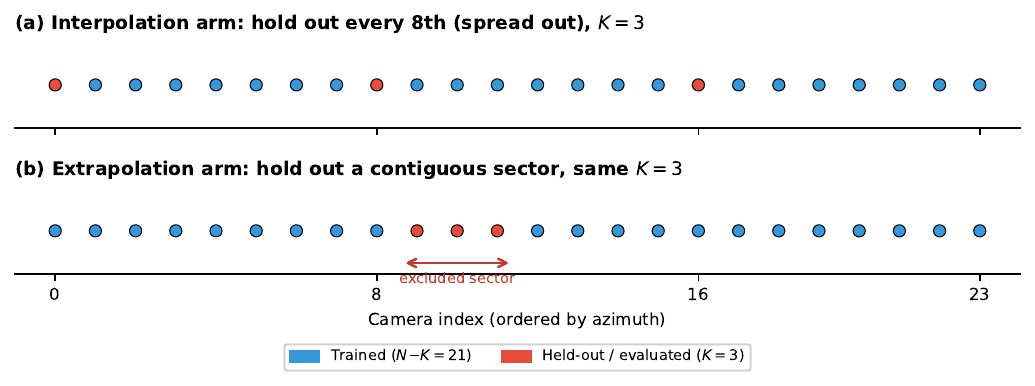}
\caption{\textbf{Interpolation vs.\ extrapolation evaluation.}
(a)~Standard evaluation: held-out frames (red) have trained
neighbors on both sides---the metric measures interpolation.
(b)~Spatial-holdout evaluation: a contiguous region is excluded
from training. Held-out frames now have no nearby training
views---the metric measures extrapolation. Our fair matched-count
protocol ensures both arms train on the same number of images and
evaluate on the same number of held-out images (though the specific
images differ).}
\label{fig:protocol}
\end{figure}

\section{Method}
\label{sec:method}

\subsection{Problem Statement}

The standard 3DGS evaluation protocol~\cite{kerbl3Dgaussians} sorts
images by filename, holds out every 8th image as the test set, and
trains on the remaining $\sim$87.5\%. Each held-out image has trained
neighbors on both sides---the metric measures how well the model
\emph{interpolates} between nearby views, not how well it synthesizes
genuinely novel viewpoints from unseen spatial regions.

\subsection{Fair Matched-Count Protocol}

We propose a controlled comparison (Figure~\ref{fig:protocol}) that
isolates the effect of holdout spatial distribution while eliminating
data-count confounds. Given a scene with $N$ cameras:

\begin{enumerate}[leftmargin=*,topsep=2pt,itemsep=1pt]
\item \textbf{Interpolation arm (standard protocol).}
  Hold out every 8th image by sorted filename ($K = \lceil N/8
  \rceil$ images). Train on $N\!-\!K$ images. Evaluate on the $K$
  held-out images. Each held-out image has trained neighbors on both
  sides---this measures \emph{interpolation}.
\item \textbf{Extrapolation arm (sector holdout).}
  Sort cameras by azimuth around the scene centroid. Starting from a
  random position (seed 42), hold out $K$ \emph{contiguous} cameras.
  Train on $N\!-\!K$ images. Evaluate on $K$ held-out
  images (different images than the interp arm, but the same count).
  Test views have substantially larger nearest-training-view
  distances---this measures \emph{extrapolation}.
\end{enumerate}

The gap is defined as:
\begin{equation}
\Delta = \text{PSNR}_\text{interp} - \text{PSNR}_\text{extrap}
\label{eq:gap}
\end{equation}

Both arms train on exactly $N\!-\!K$ images and evaluate on exactly
$K$ genuinely unseen images. The designed contrast is whether the
held-out images are spatially interleaved with training views
(interpolation) or concentrated in an unseen region (extrapolation).
By fixing the same held-out count $K$ for both arms, the comparison
isolates the effect of the holdout's spatial distribution rather than
the amount of training data.

\section{The Interpolation--Extrapolation Gap}
\label{sec:gap}

\subsection{Experimental Setup}

We evaluate the gap on two data regimes:

\paragraph{Real-capture scenes.}
Nine scenes from MipNeRF360~\cite{mipnerf360} (5~outdoor:
bicycle, flowers, garden, stump, treehill; 4~indoor: bonsai, counter,
kitchen, room) and one from Tanks and
Temples~\cite{tanksandtemples} (truck). Outdoor scenes use
\texttt{images\_4} (quarter resolution); indoor scenes use
\texttt{images\_2} (half resolution), following standard practice.

\paragraph{Generated-view scenes.}
Six scenes from the HY-World~2.0~\cite{hyworld} panorama-to-3D
pipeline: two from the released demo examples
(case000, a garden patio: 199~images;
case001, a living room: 178~images) and four generated by us
from CC0 equirectangular panoramas
(courtyard, gazebo, yaris, alps: 283~images each). All
were converted to COLMAP format for cross-method comparison.

\paragraph{Methods.}
Three published 3DGS implementations, each with default
hyperparameters and 30K training iterations:
Official 3DGS~\cite{kerbl3Dgaussians} (with gsplat~\cite{gsplat}),
Mip-Splatting~\cite{mipgs} ($\texttt{kernel\_size}\!=\!0.1$),
and 3DGS Markov chain Monte Carlo (3DGS-MCMC)~\cite{3dgsmcmc}.
To test representation-independence (\S\ref{sec:nerf}) we additionally
train a non-Gaussian volumetric NeRF, Instant-NGP~\cite{instantngp}
(hash-grid encoding, nerfacc~\cite{nerfacc} occupancy-grid rendering,
20K~steps), under the same per-arm split files.
Total: 502 training runs including multi-seed validation,
multi-sector robustness, and regularizer ablations.

\subsection{Main Results}

\paragraph{Real captures.}
Table~\ref{tab:main} shows the gap across 10 real-capture scenes.
\emph{Every} scene and \emph{every} method shows a positive gap,
ranging from \textbf{3.2 to 11.1~dB} (mean 5.84~dB for Official
3DGS). The gap is
consistent across all three metrics: SSIM shows a mean gap of 0.124
and LPIPS of 0.086 (Appendix~\ref{sec:supp_metrics}). It is consistently larger than
the inter-method performance differences on the same scenes
($<$2.5~dB on standard benchmarks).

\paragraph{Generated views.}
Table~\ref{tab:gen} shows the gap on 6~generated-view scenes. The
gap ranges from \textbf{3.3 to 12.0~dB} (mean 8.4~dB)---about
\textbf{1.4$\times$} the real-capture gap. The two HY-World example
scenes (case000, case001) show the largest gaps (10--12~dB),
while the four scenes generated from CC0 panoramas show gaps of
3.3--10.2~dB. Mip-Splatting vs.\ Official 3DGS shows 2/6 ranking
reversals (case001, gazebo). This suggests that video-diffusion-generated data is
susceptible to the same evaluation pattern as real captures, likely
because generated trajectories follow smooth paths.

\begin{table}[t]
\centering
\small
\caption{\textbf{Generated-view gap} (6 HY-World scenes, fair
matched-count protocol). Gaps are $\sim$1.4$\times$ real-capture
gaps. Numbers in parentheses are the per-column method rank
(1=best). $\dagger$Ranking reversal: the Mip-Splatting vs.\ Official
order flips between interp and extrap (single-run; not multi-seed
confirmed).}
\label{tab:gen}
\begin{tabular}{llrrr}
\toprule
Scene & Method & Interp & Extrap & Gap \\
\midrule
\multirow{3}{*}{case000}
  & Official 3DGS & 31.36~(3) & 19.40~(3) & 11.96 \\
  & Mip-Splatting & 31.68~(2) & 19.67~(2) & 12.01 \\
  & 3DGS-MCMC & \textbf{31.82}~(1) & \textbf{19.85}~(1) & 11.97 \\
\midrule
\multirow{3}{*}{case001$^\dagger$}
  & Official 3DGS & 24.34~(3) & 14.63~(2) & 9.71 \\
  & Mip-Splatting & 24.52~(2) & 14.44~(3) & 10.08 \\
  & 3DGS-MCMC & \textbf{25.67}~(1) & \textbf{14.67}~(1) & 11.00 \\
\midrule
\multirow{3}{*}{courtyard}
  & Official 3DGS & 21.83~(3) & 14.40~(3) & 7.43 \\
  & Mip-Splatting & 22.00~(2) & 14.62~(2) & 7.38 \\
  & 3DGS-MCMC & \textbf{22.45}~(1) & \textbf{14.64}~(1) & 7.80 \\
\midrule
\multirow{3}{*}{gazebo$^\dagger$}
  & Official 3DGS & 20.67~(3) & \textbf{12.50}~(1) & 8.17 \\
  & Mip-Splatting & 21.29~(2) & 11.89~(3) & 9.40 \\
  & 3DGS-MCMC & \textbf{22.60}~(1) & 12.36~(2) & 10.24 \\
\midrule
\multirow{3}{*}{yaris}
  & Official 3DGS & 17.74~(3) & 10.56~(3) & 7.18 \\
  & Mip-Splatting & 18.58~(2) & 10.82~(2) & 7.75 \\
  & 3DGS-MCMC & \textbf{18.62}~(1) & \textbf{10.90}~(1) & 7.72 \\
\midrule
\multirow{3}{*}{alps}
  & Official 3DGS & 17.34~(3) & 14.08~(3) & 3.27 \\
  & Mip-Splatting & 17.96~(2) & 14.31~(2) & 3.65 \\
  & 3DGS-MCMC & \textbf{19.04}~(1) & \textbf{14.58}~(1) & 4.46 \\
\midrule
\textbf{Mean} & Official & 22.21 & 14.26 & \textbf{7.95} \\
              & Mip-Splat & 22.67 & 14.29 & \textbf{8.38} \\
              & MCMC & \textbf{23.37} & \textbf{14.50} & \textbf{8.87} \\
\bottomrule
\end{tabular}
\end{table}

\begin{table*}[!htbp]
\centering
\small
\caption{\textbf{Fair matched-count interpolation--extrapolation gap.}
Both arms train on $N\!-\!K$ images; interp holds out every 8th
(standard protocol), extrap holds out a contiguous sector (same $K$).
All eval images are genuinely unseen.
$\star$Robust Mip-Splatting vs.\ Official reversal confirmed by
multi-seed validation.
Official 3DGS rows show mean$\pm$std over 3 training seeds;
Mip-Splatting and MCMC entries are single runs except for the
additional Mip-Splatting seed runs used to validate starred reversals.
IntR/ExtR = interpolation/extrapolation rank (1=best).}
\label{tab:main}
\begin{tabular}{l l rrr rr}
\toprule
Scene & Method & Interp & Extrap & Gap & IntR & ExtR \\
\midrule
\multirow{3}{*}{bicycle$^\star$}
  & Official 3DGS & 25.12{\tiny$\pm$.12} & \textbf{18.89}{\tiny$\pm$.15} & 6.23{\tiny$\pm$.03} & 3 & 1 \\
  & Mip-Splatting & \textbf{25.27} & 18.63 & 6.64 & 1 & 2 \\
  & 3DGS-MCMC & 25.22 & 18.53 & 6.69 & 2 & 3 \\
\midrule
\multirow{3}{*}{flowers}
  & Official 3DGS & 21.49{\tiny$\pm$.07} & 18.15{\tiny$\pm$.06} & 3.35{\tiny$\pm$.02} & 2 & 2 \\
  & Mip-Splatting & 21.36 & 18.10 & 3.27 & 3 & 3 \\
  & 3DGS-MCMC & \textbf{21.52} & \textbf{18.35} & 3.17 & 1 & 1 \\
\midrule
\multirow{3}{*}{garden$^\star$}
  & Official 3DGS & 27.29{\tiny$\pm$.05} & \textbf{23.65}{\tiny$\pm$.09} & 3.63{\tiny$\pm$.06} & 3 & 1 \\
  & Mip-Splatting & 27.37 & 23.02 & 4.35 & 2 & 3 \\
  & 3DGS-MCMC & \textbf{27.46} & 23.57 & 3.89 & 1 & 2 \\
\midrule
\multirow{3}{*}{stump}
  & Official 3DGS & 26.41{\tiny$\pm$.24} & \textbf{21.64}{\tiny$\pm$.50} & 4.77{\tiny$\pm$.26} & 2 & 1 \\
  & Mip-Splatting & 26.32 & 21.42 & 4.90 & 3 & 3 \\
  & 3DGS-MCMC & \textbf{26.72} & 21.59 & 5.13 & 1 & 2 \\
\midrule
\multirow{3}{*}{treehill}
  & Official 3DGS & 22.38{\tiny$\pm$.12} & \textbf{16.36}{\tiny$\pm$.09} & 6.03{\tiny$\pm$.04} & 2 & 1 \\
  & Mip-Splatting & 21.94 & 16.14 & 5.80 & 3 & 3 \\
  & 3DGS-MCMC & \textbf{22.40} & 16.26 & 6.14 & 1 & 2 \\
\midrule
\multirow{3}{*}{bonsai}
  & Official 3DGS & 32.07{\tiny$\pm$.06} & 24.34{\tiny$\pm$.23} & 7.74{\tiny$\pm$.19} & 2 & 2 \\
  & Mip-Splatting & 31.97 & 24.24 & 7.73 & 3 & 3 \\
  & 3DGS-MCMC & \textbf{32.44} & \textbf{25.20} & 7.24 & 1 & 1 \\
\midrule
\multirow{3}{*}{counter}
  & Official 3DGS & 29.00{\tiny$\pm$.07} & 17.91{\tiny$\pm$.15} & 11.09{\tiny$\pm$.22} & 3 & 3 \\
  & Mip-Splatting & 29.15 & 18.19 & 10.95 & 2 & 2 \\
  & 3DGS-MCMC & \textbf{29.18} & \textbf{18.31} & 10.87 & 1 & 1 \\
\midrule
\multirow{3}{*}{kitchen}
  & Official 3DGS & 31.18{\tiny$\pm$.14} & 23.91{\tiny$\pm$.12} & 7.26{\tiny$\pm$.23} & 3 & 1 \\
  & Mip-Splatting & \textbf{31.45} & 23.78 & 7.67 & 1 & 3 \\
  & 3DGS-MCMC & \textbf{31.45} & 23.79 & 7.66 & 2 & 2 \\
\midrule
\multirow{3}{*}{room}
  & Official 3DGS & 31.39{\tiny$\pm$.15} & 26.85{\tiny$\pm$.24} & 4.54{\tiny$\pm$.33} & 3 & 3 \\
  & Mip-Splatting & 31.59 & 27.13 & 4.46 & 2 & 2 \\
  & 3DGS-MCMC & \textbf{31.69} & \textbf{27.27} & 4.42 & 1 & 1 \\
\midrule
\multirow{3}{*}{truck}
  & Official 3DGS & 25.29{\tiny$\pm$.08} & 21.48{\tiny$\pm$.16} & 3.81{\tiny$\pm$.10} & 2 & 2 \\
  & Mip-Splatting & 25.26 & 21.28 & 3.98 & 3 & 3 \\
  & 3DGS-MCMC & \textbf{26.03} & \textbf{22.08} & 3.95 & 1 & 1 \\
\midrule
\textbf{Mean} & Official & 27.16 & 21.32 & \textbf{5.84} & & \\
              & Mip-Splat & 27.17 & 21.19 & \textbf{5.98} & & \\
              & MCMC & \textbf{27.41} & \textbf{21.50} & \textbf{5.92} & & \\
\bottomrule
\end{tabular}
\end{table*}

\begin{figure}[t]
\centering
\includegraphics[width=\columnwidth]{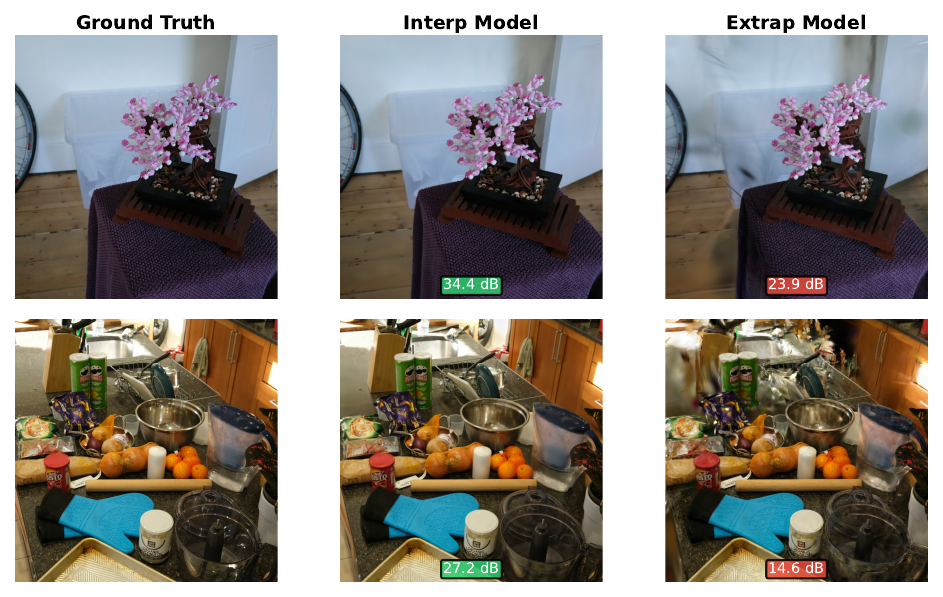}
\caption{\textbf{Same-image comparison} (Official 3DGS, shared
held-out images). Each row shows the \emph{same ground-truth (GT)
image} rendered by two models trained on different $N\!-\!K$ images.
The interp model (trained without every-8th frames) renders
faithfully; the extrap model (trained without the azimuth sector
containing this viewpoint) shows severe degradation, despite
training on the same number of images.
\emph{This is not an image-difficulty effect:} on the 31 images held
out by \emph{both} arms the gap is still $+5.81$~dB, while copying the
nearest training-view GT differs by only $+0.9$~dB between
arms~(\S\ref{sec:gap}).}
\label{fig:realcap}
\end{figure}

\paragraph{Ranking reversals.}
Multi-seed validation on both methods confirms two \emph{robust}
reversals ($\star$ in Table~\ref{tab:main}): \textbf{garden}
(Mip wins interp +0.07, Official wins extrap +0.52; 93\% of
pairwise seed combos) and \textbf{bicycle} (Mip wins interp +0.15,
Official wins extrap +0.22; 89\%). Other scenes do not show
robust reversals under multi-seed.
The consistent gap \emph{direction} (interp $>$ extrap) across all
48 scene-method combinations is the primary finding; the two
robust reversals demonstrate that the gap can be practically
consequential for method selection.

\paragraph{The gap is not just harder images.}
A natural objection is that extrapolation views are simply
intrinsically harder. Two controls rule this out.
\emph{(i)~Same-image control:} 31~images fall in \emph{both} held-out
sets (where the sector intersects every-8th positions). On these
images---identical GT and viewpoint, only the training set differs---the
gap is still +5.81~dB (all 31 positive), so it reflects model quality,
not image difficulty (Appendix~\ref{sec:supp_sameimage}).
\emph{(ii)~Copy baseline:} copying the nearest training-view GT scores
${\sim}12$~dB for \emph{both} arms (interp 12.6, extrap 11.7; a 0.9~dB
difference), an order of magnitude smaller than the 3--11~dB model
gap---so GT similarity to training views does not explain it
(Appendix~\ref{sec:supp_copybaseline}).

\section{Why Does the Gap Exist?}
\label{sec:mechanism}

\subsection{Angular Distance Drives the Gap}

\begin{figure}[t]
\centering
\includegraphics[width=\columnwidth]{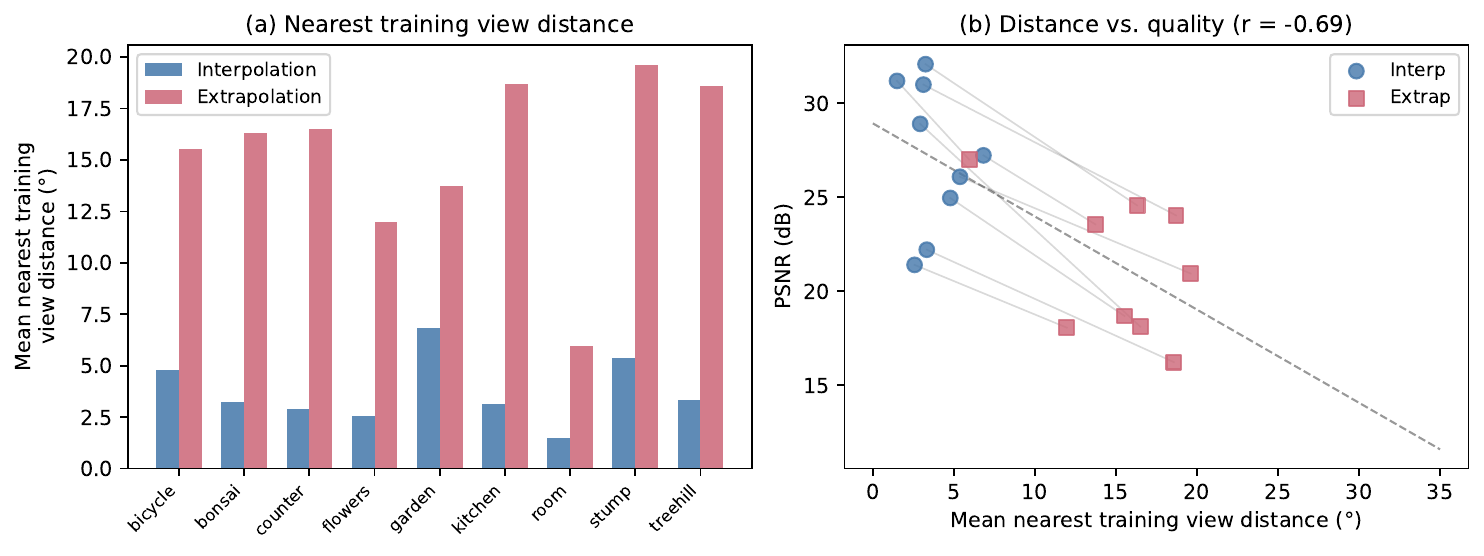}
\caption{\textbf{Nearest training-view distance and the gap.}
(a)~By construction, extrapolation test views are $4.1\times$ farther
from their nearest training view than interpolation views
(mean $15.2^\circ$ vs.\ $3.7^\circ$ across the 9 scenes).
(b)~The mechanistic test: per-image PSNR falls with nearest-view
distance ($n\!=\!492$, $r = -0.58$, $p < 10^{-44}$), and the trend
holds \emph{within} every scene. Lines connect the two arms of each
scene.}
\label{fig:angular}
\end{figure}

Figure~\ref{fig:angular} supports \emph{angular distance to training
views} as a major factor driving the gap. Across the 9~MipNeRF360
real-capture scenes (excluding truck because its Tanks-and-Temples
camera geometry differs), extrap test views are $4.1\times$ farther
from their nearest training view than interp views
(mean $15.2^\circ$ vs.\ $3.7^\circ$). This distance increase is
expected---the sector holdout raises angular distance \emph{by
construction}---so the larger separation alone is not evidence that
distance \emph{causes} the gap. The mechanistic evidence is the
per-image relationship: PSNR falls with nearest-view distance across
all 492 held-out images (Pearson $r = -0.58$, Spearman
$\rho = -0.57$, $p < 10^{-44}$), and, crucially, this trend holds
\emph{within} each scene individually (mean within-scene
$r = -0.66$, $p < 0.005$ on all 9). The within-scene correlation
removes the by-construction and scene-level confounds: quality
degrades smoothly with distance even among held-out views of the same
scene and arm. Simpler than learned uncertainty
methods~\cite{primu}, nearest-view distance is not improved on by more
complex features or nonlinear models in our data, making it an
effective zero-cost diagnostic that flags interpolation-dominated
evaluation without retraining (\S\ref{sec:discussion}).

\paragraph{Multi-sector robustness.}
To verify that the gap is not an artifact of one particular holdout
region, we rotate the sector start over 8 evenly-spaced azimuths on 4
scenes---2 outdoor (garden, bicycle) and \emph{2 indoor} (kitchen,
bonsai)---for 32 additional runs. The gap is \emph{positive at every
sector position on every scene} (Table~\ref{tab:sector}); magnitude
varies (std 1.0--2.5~dB) but direction is invariant. The two indoor
scenes---where an azimuth sector need not be spatially
contiguous---show the same invariant gap, so the effect does not
require a geometrically contiguous holdout. Main results use the
seed-42 position.

\begin{table}[t]
\centering
\small
\caption{\textbf{Multi-sector robustness} (8 sector positions per
scene, Official 3DGS). The gap is positive at every position;
magnitude varies with region difficulty.}
\label{tab:sector}
\setlength{\tabcolsep}{3.5pt}
\begin{tabular}{lrrrr}
\toprule
Scene & Gap mean & Gap std & Min gap & Max gap \\
\midrule
garden  & +2.3 & 1.0 & +0.8 & +3.6 \\
bicycle & +4.7 & 2.5 & +1.4 & +8.0 \\
kitchen & +6.1 & 1.6 & +3.7 & +9.3 \\
bonsai  & +6.6 & 1.6 & +3.7 & +9.1 \\
\bottomrule
\end{tabular}
\end{table}

\subsection{SH-Degree Decomposition of the Gap}

\begin{figure}[t]
\centering
\includegraphics[width=\columnwidth]{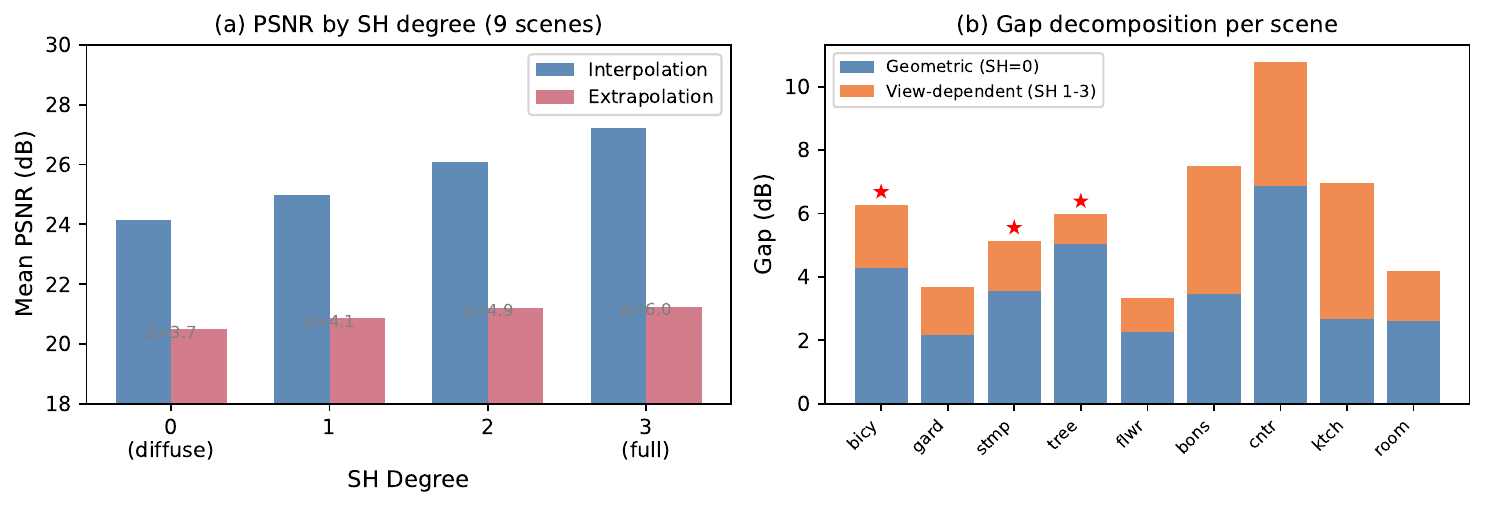}
\caption{\textbf{Gap attribution by SH degree.}
(a)~Interp PSNR improves with SH degree (view-dependent color helps);
extrap PSNR gains less. (b)~Per-scene decomposition: blue =
diffuse/geometry-proxy gap (SH=0), orange = view-dependent gap. Stars
($\star$) mark scenes where SH \emph{hurts} extrapolation. Mean: 62\%
diffuse, 38\% VD.}
\label{fig:sh}
\end{figure}

To understand \emph{why} the gap exists, we decompose it by spherical
harmonic (SH) degree---a diagnostic any future paper can apply with
zero retraining. SH coefficients above degree~0 encode
view-dependent appearance; rendering at SH degree~0 produces a
purely diffuse (view-independent) image. We render all extrap and
interp test views at SH degrees 0--3 from the same trained model
(Figure~\ref{fig:sh}).

The gap at SH degree~0 (diffuse only) captures a
\emph{diffuse/geometry-proxy} component: with view-dependent color
removed, the residual gap reflects Gaussians not positioned correctly
for novel viewpoints. The additional gap from SH degrees 1--3 captures
the \emph{view-dependent} component: SH coefficients overfit to
training-view directions. Across the same 9 MipNeRF360 scenes,
\textbf{62\%} of the gap is in the diffuse/geometry-proxy component and
\textbf{38\%} is view-dependent (Table~\ref{tab:sh}). On 3 scenes (bicycle, stump, treehill), higher
SH degrees actually \emph{hurt} extrapolation PSNR---the model's
learned view-dependent colors are wrong for novel directions,
consistent with SH overfitting to training views as documented by
concurrent work on sparse-view settings~\cite{dropansh}.
\emph{Caveat:} SH coefficients can compensate for geometric
errors~\cite{icogs} (``appearance compensation''), so the geometric
and view-dependent components interact rather than sum perfectly.
The decomposition is approximate but directionally informative.

\begin{table}[t]
\centering
\small
\caption{\textbf{SH-degree decomposition of the gap} (seed-0
models). Gap@SH0 = gap using only diffuse color (a
geometry-proxy component). VD contribution = Gap@SH3 $-$ Gap@SH0.
On 3 scenes ($\star$), higher SH \emph{hurts} extrap PSNR.
Diffuse/VD interact rather than sum exactly (see text).}
\label{tab:sh}
\setlength{\tabcolsep}{3pt}
\begin{tabular}{lrrrr}
\toprule
Scene & Gap@SH0 & Gap@SH3 & Diff.\% & VD\% \\
\midrule
bicycle$^\star$  & +4.29 & +6.27 & 68 & 32 \\
garden           & +2.17 & +3.67 & 59 & 41 \\
stump$^\star$    & +3.55 & +5.14 & 69 & 31 \\
treehill$^\star$ & +5.05 & +5.98 & 84 & 16 \\
flowers          & +2.26 & +3.34 & 68 & 32 \\
bonsai           & +3.47 & +7.51 & 46 & 54 \\
counter          & +6.87 & +10.78 & 64 & 36 \\
kitchen          & +2.67 & +6.95 & 38 & 62 \\
room             & +2.60 & +4.18 & 62 & 38 \\
\midrule
\textbf{Mean}    &       &       & \textbf{62} & \textbf{38} \\
\bottomrule
\end{tabular}
\end{table}

\subsection{The Gap Is Representation-Independent}
\label{sec:nerf}

The SH decomposition attributes most of the gap to the
diffuse/geometry-proxy component rather than Gaussian-specific
view-dependent color. If this reflects a genuine geometric cause, the
gap should appear in a representation with \emph{no} Gaussians and
\emph{no} SH at all. We test this directly with Instant-NGP, a
volumetric NeRF that encodes the scene in a multi-resolution hash grid
and a small MLP, rendered with nerfacc occupancy-grid sampling. We
train it under the identical fair matched-count protocol---the same
per-scene split files as Table~\ref{tab:main}, one model per arm,
20K~steps---on the 9 MipNeRF360 scenes (Table~\ref{tab:nerf}).

The gap reproduces on \textbf{all 9 scenes} (mean
\textbf{+3.73~dB}, 95\% CI [2.14, 5.33]; sign test $p\!=\!0.002$,
paired $t$-test $p\!=\!6\!\times\!10^{-4}$). Crucially, the per-scene
NeRF gap is strongly correlated with the 3DGS gap
($r\!=\!0.90$, $p\!=\!0.001$): counter is largest and
flowers/stump smallest under both representations, indicating a shared
scene-geometric cause rather than a representation artifact. Interp
PSNR matches published nerfacc NGP within ${\sim}1$--$2$~dB
(mean $-1.65$~dB, i.e.\ slightly conservative), confirming the
pipeline is faithful. The mean magnitude is smaller than 3DGS
(5.8~dB), consistent with NeRF's hash-grid interpolation degrading
more gracefully off-distribution than explicit primitives---but the
direction and per-scene structure are unmistakably the same.

\begin{table}[t]
\centering
\small
\caption{\textbf{The gap in a non-Gaussian NeRF.} Instant-NGP
(hash-grid + MLP, nerfacc volumetric rendering) under the same fair
matched-count protocol as Table~\ref{tab:main} (separate model per
arm, identical splits, 20K~steps), 9 MipNeRF360 scenes. All gaps
positive; per-scene gap correlates with the 3DGS-Official gap at
$r\!=\!0.90$.}
\label{tab:nerf}
\setlength{\tabcolsep}{4pt}
\begin{tabular}{lrrr}
\toprule
Scene & Interp & Extrap & Gap \\
\midrule
garden   & 24.47 & 22.08 & +2.39 \\
bicycle  & 22.33 & 19.59 & +2.74 \\
flowers  & 20.01 & 18.51 & +1.50 \\
stump    & 20.80 & 19.32 & +1.48 \\
treehill & 22.05 & 17.99 & +4.06 \\
bonsai   & 29.05 & 23.12 & +5.93 \\
counter  & 26.24 & 18.48 & +7.76 \\
kitchen  & 27.02 & 22.68 & +4.34 \\
room     & 29.70 & 26.29 & +3.41 \\
\midrule
\textbf{Mean} & \textbf{24.63} & \textbf{20.90} & \textbf{+3.73} \\
\bottomrule
\end{tabular}
\end{table}

\subsection{Regularization}
\label{sec:viewspec}

We also tested whether optimization-side interventions can close the
gap: six training-time regularizers on all 10~real-capture scenes
(Official 3DGS, extrap arm; Appendix~\ref{sec:supp_reg}). The best
(SH-degree annealing, view-dependent regularization) help only
9/10~scenes by +0.09--0.17~dB---modest against the 5.84~dB
gap---while aggressive interventions destabilize training. The gap is
largely a \emph{data-coverage} problem, not one that loss-side tuning
resolves.

\section{Implications}
\label{sec:implications}

Since loss-side tuning barely moves the gap, the practical lever is
\textbf{capturing or generating more angularly diverse training
views}, as the per-image distance correlation
(Figure~\ref{fig:angular}) predicts.

\paragraph{Coverage-guided view selection.}
If the gap is a data-coverage problem, strategically adding views
should help more than random placement. We validate this on all 10
real-capture scenes: starting from the extrap training set, we add
$M\!\in\!\{1,3,5,8\}$ images from the held-out sector, selected
either by our coverage diagnostic (greedily picking views farthest
from any training view) or randomly (5 repeats per setting; 240
total runs). Coverage-guided selection outperforms random on
\textbf{31/40} scene-budget combinations, with 16/40 significantly
above the 95\% CI of the random distribution
($t$-test, $n\!=\!5$; Table~\ref{tab:capplan}). The advantage is
strongest at $M\!=\!1$: adding a \emph{single} well-placed view
beats random on 9/10 scenes (+0.60~dB avg, 5/10 significant). \emph{Note}: coverage adds gap-filling views to training, which
also improves coverage of remaining eval images; both training
improvement and eval-set coverage contribute to the observed gains.
Unlike learned methods (FisherRF~\cite{fisherrf},
COVER~\cite{cover}) that require a trained model, our diagnostic
needs only camera poses---applicable \emph{before} any
reconstruction begins (Appendix~\ref{sec:supp_viewsel}).

\begin{table}[t]
\centering
\small
\caption{\textbf{Coverage-guided view selection} (10 scenes,
$M\!\in\!\{1,3,5,8\}$ views added, 240 runs).
$\Delta$ = coverage $-$ random mean (5 repeats).
Bold = coverage above 95\% CI of random ($p < 0.05$).}
\label{tab:capplan}
\setlength{\tabcolsep}{4pt}
\begin{tabular}{lrrrr}
\toprule
& \multicolumn{4}{c}{$\Delta$(coverage $-$ random) dB} \\
\cmidrule(lr){2-5}
Scene & $M\!=\!1$ & $M\!=\!3$ & $M\!=\!5$ & $M\!=\!8$ \\
\midrule
garden  & \textbf{+0.88} & \textbf{+0.90} & +0.71 & +0.93 \\
bicycle & +0.27 & +0.40 & \textbf{+0.87} & +0.57 \\
flowers & +0.33 & +0.25 & \textbf{+0.77} & \textbf{+0.55} \\
stump   & +0.05 & $-$0.98 & $-$2.08 & $-$0.64 \\
treehill& $-$0.04 & \textbf{+1.08} & +0.75 & $-$0.32 \\
bonsai  & \textbf{+0.80} & $-$0.03 & \textbf{+0.63} & \textbf{+1.17} \\
counter & \textbf{+1.66} & \textbf{+1.51} & +1.05 & +0.66 \\
kitchen & +0.95 & +0.94 & \textbf{+1.20} & \textbf{+0.97} \\
room    & \textbf{+0.61} & \textbf{+0.62} & +0.18 & $-$0.70 \\
truck   & \textbf{+0.54} & +0.22 & $-$0.32 & $-$0.81 \\
\midrule
\textbf{Win} & \textbf{9/10} & 8/10 & 8/10 & 6/10 \\
\textbf{Sig.} & 5/10 & 4/10 & 4/10 & 3/10 \\
\textbf{$\Delta$} & \textbf{+0.60} & +0.49 & +0.38 & +0.24 \\
\bottomrule
\end{tabular}
\end{table}

\paragraph{Downstream impact.}
The gap also propagates to downstream perception proxies: across 6
scenes, depth estimated (Depth~Anything~V2~\cite{depthanything}) from
extrap renders matches GT-photo depth worse on 5/6~scenes
(root-mean-square error, RMSE, $+$42--248\%) and no-reference
perceptual quality (MUSIQ~\cite{musiq}) drops on 6/6 (mean
$-$5.1\%). These are
proxy metrics (render-vs-GT-photo depth, no-reference quality), so we
treat them as suggestive corroboration; full results are in
Appendix~\ref{sec:supp_downstream}.

\section{Discussion}
\label{sec:discussion}

\paragraph{Implications for the field.}
Standard near-trajectory holdouts are not wrong---they faithfully
measure interpolation quality, which matters for many
applications---but they should not, on their own, be read as evidence
of spatial generalization. Because the gap exceeds the differences
typically reported between published methods
(Table~\ref{tab:main}), a comparison claiming ``0.5~dB better'' can be
smaller than the evaluation-mode effect; under multi-seed confirmation
this is consequential enough to reverse two method rankings (garden,
bicycle). We therefore recommend reporting a spatial-holdout metric
\emph{alongside} the standard one whenever generalization is claimed;
the toolkit we are preparing for release reduces this to a one-line
addition.

\paragraph{Feed-forward generality.}
We also evaluate two architecturally distinct feed-forward models that
predict Gaussians from 2 input views---MVSplat~\cite{mvsplat}
(cost-volume, ECCV'24) and DepthSplat~\cite{depthsplat} (DPT +
Depth Anything, CVPR'25)---with context views sampled from images
available to both arms (mean$\pm$std over 3 seeds; Appendix~\ref{sec:supp_mvsplat}).
Zero-shot on the 9 MipNeRF360 scenes ($256{\times}256$), the gap is
positive in \textbf{46/54} method-scene-seed combinations (sign test
$p{<}10^{-5}$; MVSplat $+0.99\pm1.00$~dB, DepthSplat
$+1.56\pm1.16$~dB), though absolute PSNR is low (${\sim}8$--$10$~dB)
due to domain mismatch. Evaluating \emph{in-domain} on 38
RealEstate10K test scenes with each method's official pipeline
(identical preprocessing; only the target frames change, from
published-nearby to farthest-from-context) removes that confound:
DepthSplat \textbf{+7.3~dB} (25.5/18.3), MVSplat \textbf{+9.1~dB}
(25.7/16.6). This feed-forward test uses a \emph{distance-based target}
protocol (nearby vs.\ farthest-from-context), an analogous stress test
of the interpolation-vs-extrapolation axis rather than our exact
azimuth-sector split. Across optimization-based Gaussians and a
volumetric NeRF (matched-count sector, \S\ref{sec:nerf}) and
feed-forward Gaussians (distance-based target), the gap appears in all
three representation families---evidence it is tied to spatial coverage,
not to any single representation or to per-scene overfitting.

\paragraph{Limitations.}
The sector holdout is one specific structured holdout; however,
the per-image correlation ($r\!=\!-0.58$) predicts that \emph{any}
holdout increasing angular distance would produce a similar gap.
The azimuth-based sector may not map to a contiguous spatial region
for indoor scenes.
Our cross-representation evaluation covers two feed-forward methods
and one volumetric NeRF (Instant-NGP) on nine scenes; broader coverage
across additional feed-forward and NeRF architectures is left to future
work.
Some ranking changes are seed-sensitive; we treat only garden and
bicycle as robust reversals.
The capture planning experiment selects from existing held-out
images rather than generating truly new viewpoints.

\paragraph{Training non-determinism.}
3DGS training is non-deterministic (\texttt{atomicAdd} in the
backward pass). Across 3 seeds on all 10 scenes (Official 3DGS), the
mean gap is 39$\times$ the mean seed std (worst case room,
$13\times$), and the direction is positive on all 30 scene-seed
combinations (Appendix~\ref{sec:supp_multiseed}).

\paragraph{Recommended evaluation practice.}
We recommend future 3DGS papers report: (1)~standard
\texttt{test\_every=N} for comparability; (2)~a spatial-holdout metric
(one extra training run); (3)~the holdout construction rule; and
(4)~nearest-training-view angular distance per test view---a zero-cost
diagnostic that flags interpolation-dominated evaluation even without
retraining.

\paragraph{Reproducibility.}
The supplement documents every split-construction rule, per-scene
configuration, and baseline commit hash, and an artifact manifest
(Appendix~\ref{sec:supp_manifest}) maps every table and figure to its generating
procedure and source data; code is in preparation for release
(\S\ref{sec:intro}).

\section{Conclusion}
\label{sec:conclusion}

Standard MipNeRF360-style 3DGS evaluation primarily measures
near-trajectory interpolation. Our fair matched-count protocol
isolates a consistent \textbf{3--12~dB} gap---larger than inter-method
differences, robust to training noise, and reproduced across three
representation families including a non-Gaussian NeRF. It should be
reported via a spatial-holdout metric \emph{alongside} the standard
one when generalization is claimed; our toolkit makes this practical.


\clearpage
\appendix
\section*{Appendix: Supplementary Material}
\noindent This appendix provides multi-metric results, multi-seed
validation details, method configurations, and an artifact manifest.
All results use the \textbf{fair matched-count protocol} (both arms
train on $N\!-\!K$ images, evaluate on $K$ genuinely unseen images).

\section{Multi-Metric Results (SSIM and LPIPS)}
\label{sec:supp_metrics}

Table~\ref{tab:supp_metrics_real} reports the SSIM and LPIPS gap
alongside PSNR for real-capture scenes (Official 3DGS, seed~0).
The gap is consistent across all three metrics on every scene.

\begin{table}[h]
\centering
\small
\caption{Multi-metric gap on 10 real-capture scenes (Official 3DGS,
fair matched-count protocol, single seed).
$\Delta$PSNR/SSIM = interp $-$ extrap (positive = interp better).
LPIPS is shown raw (lower = better).}
\label{tab:supp_metrics_real}
\setlength{\tabcolsep}{3pt}
\begin{tabular}{lrrrrrrr}
\toprule
& \multicolumn{3}{c}{PSNR (dB)} & \multicolumn{2}{c}{SSIM} & \multicolumn{2}{c}{LPIPS} \\
\cmidrule(lr){2-4} \cmidrule(lr){5-6} \cmidrule(lr){7-8}
Scene & Int & Ext & $\Delta$ & Int & Ext & Int & Ext \\
\midrule
bicycle   & 24.96 & 18.68 & +6.27 & .754 & .551 & .216 & .334 \\
bonsai    & 32.07 & 24.56 & +7.51 & .941 & .883 & .207 & .323 \\
counter   & 28.90 & 18.12 & +10.78 & .907 & .708 & .203 & .347 \\
flowers   & 21.40 & 18.06 & +3.34 & .595 & .486 & .341 & .388 \\
garden    & 27.22 & 23.55 & +3.67 & .863 & .779 & .109 & .170 \\
kitchen   & 30.98 & 24.03 & +6.95 & .925 & .821 & .130 & .214 \\
room      & 31.18 & 27.00 & +4.18 & .917 & .872 & .224 & .257 \\
stump     & 26.08 & 20.94 & +5.14 & .751 & .550 & .227 & .348 \\
treehill  & 22.21 & 16.23 & +5.98 & .624 & .464 & .329 & .398 \\
truck     & 25.17 & 21.31 & +3.87 & .878 & .806 & .150 & .221 \\
\midrule
\textbf{Mean} & 27.02 & 21.25 & \textbf{+5.77} & .815 & .692 & .214 & .300 \\
\bottomrule
\end{tabular}
\end{table}

\section{Multi-Seed Validation}
\label{sec:supp_multiseed}

The main real-capture table reports Official 3DGS results as
mean$\pm$std over 3 training seeds (seeds 0, 1, 2) on all 10
real-capture scenes. Each scene-method pair trains \emph{two
separate models} (one per arm), so training noise is independent
between arms---it does \emph{not} cancel. Nevertheless, the gap
is robust:

\begin{itemize}[leftmargin=*,itemsep=2pt]
\item The gap is positive on all 30 scene-seed combinations.
\item Seed-to-seed gap std ranges from 0.02~dB (flowers) to
  0.33~dB (room), while the gap itself ranges from 3.4 to 11.1~dB.
\item The mean gap (5.84~dB) is 39$\times$ larger than the mean
  seed std (0.15~dB).
\item Even the worst-case scene (room: 4.54$\pm$0.33~dB) has a
  gap/noise ratio of 13.6$\times$.
\end{itemize}

Mip-Splatting has multi-seed validation (2--5 seeds) on the two
reversal scenes (garden, bicycle); other Mip and all MCMC rows
are single runs. The
consistent gap direction across all 48 scene-method combinations
(16 scenes $\times$ 3 methods, including the single-run methods)
provides additional breadth-based robustness, although the
scene-method combinations are not statistically independent.

\section{Method Configurations}
\label{sec:supp_configs}

Optimization-based methods trained with 30K iterations on the same
data splits. Total: 502 optimization-based training runs (96 main
$+$ 40 Official multi-seed $+$ 40 Mip multi-seed $+$ 32 multi-sector
$+$ 54 regularizers $+$ 240 capture planning) plus 130 feed-forward
inference evaluations (54 cross-domain on MipNeRF360 $+$ 76
in-domain on RealEstate10K).

\begin{table}[h]
\centering
\small
\caption{Training configurations for cross-method comparison.}
\label{tab:supp_configs}
\begin{tabular}{lll}
\toprule
Parameter & Official 3DGS & Mip-Splatting \\
\midrule
Iterations & 30,000 & 30,000 \\
SH degree & 3 & 3 \\
Resolution & default ($-1$) & default ($-1$) \\
Densification & standard & standard \\
kernel\_size & --- & 0.1 \\
Seeds & 0, 1, 2 & 0 (main); 1--2+ for reversals \\
\midrule
 & \multicolumn{2}{c}{3DGS-MCMC} \\
\midrule
cap\_max & \multicolumn{2}{c}{scene-dependent (1.3--5.9M)} \\
noise\_lr & \multicolumn{2}{c}{5e5} \\
scale\_reg & \multicolumn{2}{c}{0.01} \\
opacity\_reg & \multicolumn{2}{c}{0.01} \\
init\_type & \multicolumn{2}{c}{sfm} \\
seed & \multicolumn{2}{c}{42} \\
\bottomrule
\end{tabular}
\end{table}

\section{Artifact Manifest}
\label{sec:supp_manifest}

Every number in the main text traces to tracked result summaries or
per-run JSON files. Real-capture split JSONs are stored alongside
their per-scene result folders. The public benchmark summary is
\texttt{lto\_benchmark/baselines.json}; the training-command manifest
is stored under \texttt{results/fair\_lto/}. In the manifest below,
\texttt{R} abbreviates \texttt{results/fair\_lto}.

{\small
\begin{itemize}[leftmargin=*,itemsep=2pt]
\item \textbf{Generated-view table}:
  the benchmark summary stores the released generated-scene baseline
  results; per-run outputs use method/split subdirectories when the
  generated result bundle is present.
\item \textbf{Real-capture table}:
  per-scene folders contain Official 3DGS, Mip-Splatting, and MCMC
  method/split subdirectories with \texttt{lto\_eval.json} summaries.
\item \textbf{Multi-sector table}:
  \texttt{R/multi\_sector/\{scene\}/} \\
  \texttt{sector\_*/eval.json}.
\item \textbf{SH-decomposition table}:
  \texttt{R/sh\_decomposition/\{scene\}.json}.
\item \textbf{Regularizer table}:
  \texttt{R/\{scene\}/reg\_\{name\}\_extrap/} \\
  \texttt{lto\_eval.json}. Launcher commands are generated by
  \texttt{scripts/launch\_all\_tracks.sh} and
  \texttt{scripts/run\_exp\_round2.sh}.
\item \textbf{Downstream table}:
  \texttt{R/downstream\_depth/\{scene\}/} \\
  \texttt{depth\_results.json};
  \texttt{R/downstream\_impact/} \\
  \texttt{\{scene\}/results.json}.
\item \textbf{Same-image control}:
  \texttt{R/same\_image\_control/} \\
  \texttt{summary.json}, with per-image inputs sourced from
  \texttt{R/sh\_decomposition/\{scene\}.json}.
\item \textbf{Feed-forward (MVSplat + DepthSplat)}:
  \texttt{R/\{scene\}/mvsplat\_seed\{42,123,7\}.json} and
  \texttt{R/\{scene\}/depthsplat\_seed\{42,123,7\}.json}
  (9 scenes $\times$ 3 seeds $\times$ 2 methods = 54 files).
  Inference scripts:
  \texttt{scripts/eval\_mvsplat\_feedforward.py},
  \texttt{scripts/eval\_depthsplat\_feedforward.py}.
  Sources (git submodules):
  \texttt{third\_party/mvsplat/},
  \texttt{third\_party/depthsplat/}.
\item \textbf{NGP NeRF table}:
  \texttt{R/\{scene\}/} \\
  \texttt{nerf\_ngp\_\{interp,extrap\}.json}
  (9 scenes $\times$ 2 arms = 18 files). Training script:
  \texttt{scripts/train\_ngp\_nerf\_lto.py}; build and patch notes in
  \texttt{third\_party/README.md}.
\item \textbf{Nearest-image copy baseline}:
  \texttt{R/nearest\_image\_baseline/summary.json}
  (per-scene PSNR and SSIM for both splits).
\item \textbf{Capture-planning table}:
  \texttt{R/capture\_planning/} (coverage vs.\ random, $M\!\in\!\{1,3,5,8\}$,
  5 repeats, 240 runs).
\item \textbf{Feed-forward in-domain (RealEstate10K)}:
  \texttt{R/re10k\_indomain/} \\
  \texttt{\{mvsplat,depthsplat\}\_re10k.json} and the modified target
  index \texttt{evaluation\_index\_re10k\_extrap.json}.
\item \textbf{Figures}: generated by
  \texttt{scripts/generate\_figures.py} (quantitative panels) and
  \texttt{scripts/render\_qualitative.py} (same-image qualitative
  comparison), reading the result JSONs above.
\end{itemize}
}
The released toolkit scripts are under \texttt{lto\_benchmark/}; the
larger experiment launchers under \texttt{scripts/} use the
repo-relative path configuration in \texttt{scripts/env.sh}.
The released bundle uses repo-relative paths throughout; absolute
machine paths are scrubbed before packaging.

\paragraph{Baseline commit hashes.}
For exact reproducibility, baselines were run at fixed upstream
commits: Official 3DGS \texttt{b9da16c}, Mip-Splatting
\texttt{dda02ab}, 3DGS-MCMC \texttt{7b4fc9f}; NGP NeRF uses nerfacc
with Instant-NGP hash encoding (tiny-cuda-nn), with build details in
\texttt{third\_party/README.md}.

\section{Spatial-Holdout Benchmark Toolkit}
\label{sec:supp_toolkit}

We release a \emph{spatial-holdout benchmark toolkit} with
standardized splits, evaluation scripts, and baseline results. The
repository keeps the historical \texttt{lto\_benchmark/} directory name
for compatibility with the launch scripts. Usage:

{\scriptsize
\begin{verbatim}
# 1. Create sector holdout split
python lto_benchmark/create_sector_holdout.py \
  --source_path /path/to/colmap_scene \
  --output split_extrap.json

# 2. Train with held-out sector removed
python train.py -s data_ext -m model_ext \
  --iterations 30000

# 3. Evaluate on held-out views
python lto_benchmark/evaluate_lto.py \
  --source_path /path/to/full_scene \
  --model_path model_ext \
  --split split_extrap.json

# 4. Coverage diagnostic
python lto_benchmark/coverage_diagnostic.py \
  --source_path /path/to/colmap_scene
\end{verbatim}
}

Output: \texttt{lto\_eval.json} with PSNR, SSIM, LPIPS per image.
The coverage diagnostic reports mean nearest-neighbor distance,
largest angular gap, and recommended view positions---a zero-cost
check for interpolation-dominated evaluation.

Contents: \texttt{lto\_benchmark/} contains
\texttt{create\_sector\_holdout.py}, \texttt{evaluate\_lto.py},
\texttt{coverage\_diagnostic.py}, and \texttt{baselines.json}
(16 scenes $\times$ 3 methods).

\section{View Selection Method Comparison}
\label{sec:supp_viewsel}

Table~\ref{tab:supp_comparison} compares our coverage-guided view
selection against existing learned approaches. Our method's key
advantage is \emph{pre-training applicability}: it uses only camera
poses (available from COLMAP before any 3DGS training), enabling
capture planning at the data-collection stage rather than the
reconstruction stage. This is important for practical workflows
where a practitioner must decide where to place cameras \emph{before}
investing in expensive 3DGS/NeRF training.

\begin{table}[h]
\centering
\small
\caption{Comparison of view selection methods. Our coverage
diagnostic is the only zero-cost, pre-training-applicable approach.}
\label{tab:supp_comparison}
\setlength{\tabcolsep}{2.5pt}
\begin{tabular}{lcccc}
\toprule
& Trained & GPU & Pre-train & Quality \\
Method & model? & cost & applicable? & signal \\
\midrule
FisherRF~\cite{fisherrf} & Yes & High & No & Fisher info \\
COVER~\cite{cover} & Yes & Medium & No & Projection \\
PRIMU~\cite{primu} & Yes & Medium & No & Learned feat. \\
\textbf{Ours} & \textbf{No} & \textbf{Zero} & \textbf{Yes} & Angular dist. \\
\bottomrule
\end{tabular}
\end{table}

The practical importance: in robotics path planning, real-estate
scanning, autonomous driving data collection, and world-model
training pipelines, the camera placement decision must be made
\emph{before} any 3D reconstruction. Learned methods cannot help
at this stage because they require an already-trained model. Our
coverage diagnostic fills this gap by providing actionable guidance
from camera poses alone, validated by the capture planning
experiment (Table~\ref{tab:capplan}) where coverage-guided selection
outperforms random on 31/40 scene-budget combinations
(10 scenes, 240 runs).

\section{Same-Image Control}
\label{sec:supp_sameimage}

A potential confound is that the interp and extrap arms evaluate
mostly different images. To rule out image-difficulty effects, we
identify images appearing in \emph{both} eval sets (2--5 per scene,
31 total across 9 MipNeRF360 scenes). On these shared images, both
the interp model and the extrap model are evaluated against the
\emph{same GT image}---the only difference is which model rendered
it. The gap on shared images is +5.81~dB (comparable to the overall mean gap of 5.84~dB),
with all 31 positive (Table~\ref{tab:supp_sameimage}).

\begin{table}[h]
\centering
\small
\caption{Same-image control: gap on images in both eval sets.}
\label{tab:supp_sameimage}
\begin{tabular}{lrrr}
\toprule
Scene & Shared & Mean gap (dB) & All pos? \\
\midrule
garden   & 3 & +3.92 & Yes \\
bicycle  & 2 & +1.84 & Yes \\
flowers  & 4 & +2.12 & Yes \\
stump    & 2 & +2.90 & Yes \\
treehill & 3 & +2.97 & Yes \\
bonsai   & 5 & +8.35 & Yes \\
counter  & 4 & +9.40 & Yes \\
kitchen  & 4 & +8.19 & Yes \\
room     & 4 & +7.35 & Yes \\
\midrule
\textbf{Total} & \textbf{31} & \textbf{+5.81} & \textbf{Yes} \\
\bottomrule
\end{tabular}
\end{table}

Result JSON: \texttt{summary.json} under
\texttt{R/same\_image\_control/}. Per-image data comes from
\texttt{\{scene\}.json} under \texttt{R/sh\_decomposition/}.

\section{Nearest-Image Copy Baseline}
\label{sec:supp_copybaseline}

Table~\ref{tab:supp_copybaseline} reports per-scene PSNR and SSIM
for the nearest-training-GT copy baseline. For each held-out image,
we find the nearest training image by angular distance and compute
metrics between the two GT photographs. The small copy gap
(0.9~dB PSNR, 0.013 SSIM) compared to the model gap (3--11~dB)
confirms that the model gap is not explained by GT image similarity
alone.

\begin{table}[h]
\centering
\small
\caption{Nearest-image copy baseline: PSNR and SSIM between each
held-out GT and its nearest training GT. The copy gap is several-fold
to an order of magnitude smaller than the model gap.}
\label{tab:supp_copybaseline}
\setlength{\tabcolsep}{3pt}
\begin{tabular}{lrrrrrr}
\toprule
& \multicolumn{3}{c}{Copy PSNR (dB)} & \multicolumn{3}{c}{Copy SSIM} \\
\cmidrule(lr){2-4} \cmidrule(lr){5-7}
Scene & Int & Ext & $\Delta$ & Int & Ext & $\Delta$ \\
\midrule
garden   & 14.6 & 13.8 & +0.8 & .144 & .144 & .000 \\
bicycle  & 11.3 & 11.8 & $-$0.5 & .112 & .124 & $-$.012 \\
flowers  &  9.0 &  9.5 & $-$0.5 & .082 & .096 & $-$.014 \\
stump    & 13.2 & 12.7 & +0.5 & .146 & .125 & +.021 \\
treehill &  9.3 & 11.0 & $-$1.7 & .124 & .128 & $-$.004 \\
bonsai   & 15.1 & 13.6 & +1.5 & .445 & .538 & $-$.093 \\
counter  & 13.6 & 10.2 & +3.4 & .391 & .274 & +.117 \\
kitchen  & 15.2 & 12.4 & +2.8 & .296 & .249 & +.047 \\
room     & 12.1 & 10.5 & +1.6 & .405 & .354 & +.051 \\
\midrule
\textbf{Mean} & 12.6 & 11.7 & \textbf{+0.9} & .238 & .226 & \textbf{+.013} \\
\bottomrule
\end{tabular}
\end{table}

\section{Feed-Forward Generality: Per-Scene Results}
\label{sec:supp_mvsplat}

Tables~\ref{tab:supp_mvsplat} and~\ref{tab:supp_depthsplat} report
per-scene results for two architecturally distinct feed-forward models.
For each scene, 2 context views are selected from images available to
both interp and extrap arms via farthest-point sampling; 3
context-selection seeds vary the initialization.
All evaluations use $256{\times}256$ with aspect-ratio-preserving
rescale and center crop, matching each model's training pipeline.

\textbf{MVSplat}~\cite{mvsplat} (ECCV'24 Oral) uses a cost-volume
encoder with a UniMatch backbone; we use \texttt{re10k.ckpt}.
\textbf{DepthSplat}~\cite{depthsplat} (CVPR'25) uses a DPT head
with a Depth Anything V2 / DINOv2 backbone; we use the base
256$\times$256 2-view checkpoint.
The xformers dependency was replaced with PyTorch's native
\texttt{scaled\_dot\_product\_attention} for hardware portability
(verified to produce identical PSNR).

\begin{table}[h]
\centering
\small
\caption{MVSplat feed-forward evaluation. Gap = interp $-$ extrap PSNR (dB).}
\label{tab:supp_mvsplat}
\setlength{\tabcolsep}{3pt}
\begin{tabular}{lrrrrl}
\toprule
Scene & Seed 42 & Seed 123 & Seed 7 & Mean$\pm$Std & 3/3? \\
\midrule
garden   & $+$1.00 & $-$0.01 & $-$0.45 & $+$0.18$\pm$0.61 & \\
bicycle  & $+$1.67 & $+$1.00 & $+$1.43 & $+$1.37$\pm$0.28 & \checkmark \\
flowers  & $-$0.14 & $+$0.51 & $+$0.76 & $+$0.38$\pm$0.38 & \\
stump    & $+$2.35 & $+$2.45 & $+$2.57 & $+$2.46$\pm$0.09 & \checkmark \\
treehill & $+$0.40 & $+$0.92 & $+$1.00 & $+$0.77$\pm$0.27 & \checkmark \\
bonsai   & $+$2.40 & $+$2.93 & $+$2.34 & $+$2.56$\pm$0.27 & \checkmark \\
counter  & $+$0.99 & $+$1.07 & $+$0.98 & $+$1.01$\pm$0.04 & \checkmark \\
kitchen  & $-$0.99 & $-$0.48 & $+$0.16 & $-$0.44$\pm$0.47 & \\
room     & $+$0.97 & $+$0.47 & $+$0.46 & $+$0.63$\pm$0.24 & \checkmark \\
\midrule
\multicolumn{4}{l}{\textbf{Overall}: 22/27 positive ($p{<}0.002$)} & \textbf{$+$0.99$\pm$1.00} & 6/9 \\
\bottomrule
\end{tabular}
\end{table}

\begin{table}[h]
\centering
\small
\caption{DepthSplat feed-forward evaluation. Gap = interp $-$ extrap PSNR (dB).}
\label{tab:supp_depthsplat}
\setlength{\tabcolsep}{3pt}
\begin{tabular}{lrrrrl}
\toprule
Scene & Seed 42 & Seed 123 & Seed 7 & Mean$\pm$Std & 3/3? \\
\midrule
garden   & $+$1.24 & $+$2.11 & $+$1.17 & $+$1.51$\pm$0.43 & \checkmark \\
bicycle  & $+$0.73 & $+$2.64 & $+$1.96 & $+$1.78$\pm$0.79 & \checkmark \\
flowers  & $+$0.55 & $-$0.12 & $+$1.36 & $+$0.60$\pm$0.61 & \\
stump    & $+$2.73 & $+$0.50 & $+$2.71 & $+$1.98$\pm$1.05 & \checkmark \\
treehill & $+$1.33 & $+$1.41 & $+$2.84 & $+$1.86$\pm$0.69 & \checkmark \\
bonsai   & $+$3.92 & $+$3.74 & $+$3.70 & $+$3.79$\pm$0.10 & \checkmark \\
counter  & $+$1.74 & $+$1.44 & $+$1.54 & $+$1.57$\pm$0.12 & \checkmark \\
kitchen  & $+$1.24 & $+$1.14 & $+$0.82 & $+$1.07$\pm$0.18 & \checkmark \\
room     & $-$0.17 & $+$0.27 & $-$0.45 & $-$0.12$\pm$0.30 & \\
\midrule
\multicolumn{4}{l}{\textbf{Overall}: 24/27 positive ($p{<}0.0001$)} & \textbf{$+$1.56$\pm$1.16} & 7/9 \\
\bottomrule
\end{tabular}
\end{table}

\section{Downstream Impact}
\label{sec:supp_downstream}

We test whether the gap propagates to downstream perception proxies on
6 MipNeRF360 scenes where depth and perceptual proxies were available
(Official 3DGS, fair protocol; Table~\ref{tab:downstream}).

\noindent\emph{Depth estimation.}
We run Depth Anything V2~\cite{depthanything} on rendered images and
measure how well the estimated depth matches depth estimated from the
corresponding GT photograph (Pearson correlation and RMSE). On
\textbf{5/6 scenes}, extrap renders produce less consistent depth
(RMSE +42--248\%). This is a proxy for geometric fidelity (comparing
depth from renders vs.\ from GT photos, not metric depth).

\noindent\emph{Perceptual quality.}
No-reference perceptual quality (MUSIQ~\cite{musiq}) drops on all
6/6~scenes (mean $-$5.1\%), suggesting perceptual degradation
consistent with the PSNR gap. Both are proxy-on-proxy measurements and
should be read as suggestive rather than definitive.

\begin{table}[h]
\centering
\small
\caption{\textbf{Downstream impact} on 6 real scenes (Official 3DGS).
Depth: RMSE between depth(render) and depth(GT photo); lower = better.
MUSIQ: no-reference perceptual quality; higher = better.
$\dagger$Room is the only scene where extrap depth RMSE improves.}
\label{tab:downstream}
\setlength{\tabcolsep}{3pt}
\begin{tabular}{lrrrrr}
\toprule
& \multicolumn{3}{c}{Depth RMSE ($\downarrow$)} & \multicolumn{2}{c}{MUSIQ ($\uparrow$)} \\
\cmidrule(lr){2-4} \cmidrule(lr){5-6}
Scene & Interp & Extrap & $\Delta$\% & Interp & Extrap \\
\midrule
bicycle & .022 & .042 & +92 & 73.8 & 69.4 \\
bonsai  & .024 & .040 & +69 & 62.5 & 61.4 \\
counter & .021 & .074 & +248 & 61.1 & 58.5 \\
garden  & .012 & .018 & +42 & 75.7 & 74.2 \\
kitchen & .014 & .025 & +75 & 68.6 & 62.7 \\
room$^\dagger$ & .049 & .035 & $-$28 & 55.0 & 50.6 \\
\midrule
\textbf{Mean} & .024 & .039 & +83 & 66.1 & 62.8 \\
\bottomrule
\end{tabular}
\end{table}

\section{Regularization Ablation}
\label{sec:supp_reg}

To test whether optimization-side interventions can improve
spatial-holdout quality, we evaluate six training-time regularizers on
all 10~real-capture scenes (Official 3DGS, extrap arm, 30K steps;
Table~\ref{tab:reg}).
\textbf{SH degree annealing} (delaying SH degree increases to
7.5K/15K/22.5K) improves \textbf{9/10}~scenes (+0.17~dB mean).
\textbf{View-dependent regularization} (penalizing color variation
under small camera perturbations) also improves 9/10. \textbf{Scale
regularization} improves 8/10. All three outperform Gaussian
dropout~\cite{dropoutgs,dropgaussian} (2/10, $-$0.39~dB) and
camera jitter (2/10, $-$2.85~dB). Coverage-Aware Training (CAT),
a purpose-designed aggressive intervention, worsens all 9~tested
scenes ($-$3.1~dB), demonstrating that aggressive regularization
destabilizes training. The mild regularizers' improvements
(+0.09--0.17~dB) are modest compared to the gap itself (5.84~dB),
reinforcing that the gap is largely a data-coverage problem.

\begin{table}[h]
\centering
\small
\caption{\textbf{Regularizer comparison on extrapolation PSNR}
(10 scenes, Official 3DGS, 30K steps). $\uparrow$ = scenes improved
vs.\ multi-seed baseline. SH annealing and view-dependent
regularization each improve 9/10 scenes.}
\label{tab:reg}
\setlength{\tabcolsep}{4pt}
\begin{tabular}{lrrr}
\toprule
Method & Mean $\Delta$ (dB) & $\uparrow$ & Note \\
\midrule
Dropout 30\% & $-$0.39 & 2/10 & Hurts most scenes \\
Cam jitter $2^\circ$ & $-$2.85 & 2/10 & Too aggressive \\
SH anneal & \textbf{+0.17} & \textbf{9/10} & Best mean gain \\
ViewDep reg & +0.12 & \textbf{9/10} & Consistent \\
Scale reg & +0.09 & 8/10 & Consistent \\
CAT$^\dagger$ & $-$3.10 & 0/9 & Destabilizes \\
\bottomrule
\end{tabular}

{\footnotesize $\dagger$Coverage-Aware Training (SH reset $+$ virtual
camera $10^\circ$ $+$ opacity entropy); tested on 9 scenes.}
\end{table}

\section{Protocol Audit of Recent 3DGS Papers}
\label{sec:supp_audit}

Table~\ref{tab:supp_audit} surveys 11 recent 3DGS papers. The split
protocol for each entry was verified from the paper's released source
code (training scripts or config files) or, where code was unavailable,
from the paper's methodology section describing the evaluation split.
All use \texttt{llffhold=8} (every 8th image as test) on
smooth-trajectory datasets, making all vulnerable to the
interpolation-extrapolation gap documented in the main text.

\begin{table}[h]
\centering
\scriptsize
\caption{Evaluation protocol audit. All 11 papers use the same
near-trajectory holdout on smooth captures. Provenance: how
the split protocol was verified.}
\label{tab:supp_audit}
\setlength{\tabcolsep}{2pt}
\begin{tabular}{lllll}
\toprule
Paper & Venue & Dataset & Split & Provenance \\
\midrule
3DGS~\cite{kerbl3Dgaussians} & TOG'23 & MipNeRF360 & every 8th & code \\
Mip-Splatting~\cite{mipgs} & CVPR'24 & MipNeRF360 & every 8th & code \\
3DGS-MCMC~\cite{3dgsmcmc} & NeurIPS'24 & MipNeRF360 & every 8th & code \\
2DGS~\cite{2dgs} & SIGGRAPH'24 & MipNeRF360 & every 8th & code \\
Scaffold-GS~\cite{scaffoldgs} & CVPR'24 & MipNeRF360 & every 8th & code \\
Compact3D~\cite{compact3dgs} & CVPR'24 & MipNeRF360 & every 8th & code \\
AbsGS~\cite{absgs} & ACMMM'24 & MipNeRF360 & every 8th & code \\
DropGaussian~\cite{dropgaussian} & CVPR'25 & MipNeRF360 & every 8th & code \\
DropoutGS~\cite{dropoutgs} & CVPR'25 & MipNeRF360 & every 8th & code \\
EDGS~\cite{edgs} & CVPR'26 & MipNeRF360 & every 8th & code \\
DropAnSH~\cite{dropansh} & CVPR'26 & MipNeRF360 & every 8th & paper \\
\bottomrule
\end{tabular}
\end{table}

\end{document}